\documentclass{article}

\usepackage{arxiv}

\usepackage[utf8]{inputenc} 
\usepackage[T1]{fontenc}    
\usepackage{hyperref}       
\hypersetup{
    colorlinks=true,
    citecolor={blue!100!black},
    linkcolor={blue!100!black},
    urlcolor={blue!80!black}
}

\usepackage{url}            
\usepackage{booktabs}       
\usepackage{amsfonts}       
\usepackage{nicefrac}       
\usepackage{microtype}      
\usepackage{cite} 
\usepackage{graphicx}

\usepackage{amsmath}
\usepackage{gensymb} 
\usepackage{adjustbox}
\usepackage{svg}
\usepackage{lipsum}

\graphicspath{ {./images/} }

\title{Hitting the Gym: Reinforcement Learning Control of Exercise-Strengthened Biohybrid Robots in Simulation}

\author{
 Saul Schaffer\ \\
  Mechanical Engineering Department \\
  Carnegie Mellon University\\
   \And
 Hima Hrithik Pamu \\
  Mechanical Engineering Department \\
  Carnegie Mellon University\\
  \And
   Victoria A. Webster-Wood \\
    Mechanical Engineering Department \\
    Biomedical Engineering Department \\
    McGowan Institute for Regenerative Medicine \\
    Robotics Institute \\
    Carnegie Mellon University\\
  \texttt{vwebster@andrew.cmu.edu}
}

\begin{document}

\maketitle

\begin{abstract}

Animals can accomplish many incredible behavioral feats across a wide range of operational environments and scales that current robots struggle to match. One explanation for this performance gap is the extraordinary properties of the biological materials that comprise animals, such as muscle tissue. Using living muscle tissue as an actuator can endow robotic systems with highly desirable properties such as self-healing, compliance, and biocompatibility. Unlike traditional soft robotic actuators, living muscle biohybrid actuators exhibit unique adaptability, growing stronger with use. The dependency of a muscle's force output on its use history endows muscular organisms the ability to dynamically adapt to their environment, getting better at tasks over time. While muscle adaptability is a benefit to muscular organisms, it currently presents a challenge for biohybrid researchers: how does one design and control a robot whose actuators' force output changes over time? Here, we incorporate muscle adaptability into a many-muscle biohybrid robot design and modeling tool, leveraging reinforcement learning as both a co-design partner and system controller. As a controller, our learning agents coordinated the independent contraction of 42 muscles distributed on a lattice worm structure to successfully steer it towards eight distinct targets while incorporating muscle adaptability.  As a co-design tool, our agents enable users to identify which muscles are important to accomplishing a given task. Our results show that adaptive agents outperform non-adaptive agents in terms of maximum rewards and training time. Together, these contributions can both enable the elucidation of muscle actuator adaptation and inform the design and modeling of adaptive, performant, many-muscle robots. 

\end{abstract}


\section{Introduction}

One challenge in the design and deployment of biohybrid robots is that their actuators change their force output over time, getting stronger with exercise. This adaptation through exercise is integral to the successful function of organisms. Organisms serve as the inspiration for biohybrid robots \cite{raman2024biofabrication, raman2024soft}, with biohybrid researchers recapitulating some of their impressive capabilities, including self-healing \cite{Raman2017} biocompatibility \cite{Ricotti2017BiohybridCells} and compliance \cite{Webster-Wood2017, morimoto2020biohybrid}. However, exercise-induced adaptability makes designing and controlling biohybrid robots difficult, as the force output of the actuators is transient. As a result, creating and controlling longitudinally operational many-actuator biohybrid robots remains a challenge. It takes an incredible amount of resources to create even a relatively simple biohybrid robot crawler \cite{Raman2017b}, including labor costs, graduate student time, culture plastic, culture media, and capital equipment. These resource needs are compounded if the system needs to be purpose-built and meet operational performance standards. Without adequate modeling, building biohybrid robots can easily become an inefficient, trial-and-error (Edisonian) endeavor. However, to model a biohybrid robot, one needs to capture the mechanics, actuation, and control of that system, as they are deeply coupled in biohybrid robots \cite{Ricotti2017BiohybridCells}.

Biohybrid muscle actuation is commonly modeled as an immutable force output, independent of the use history of that actuator. However, the literature suggests that muscle adaptation in response to use history is a major contributor to expected muscle force output \cite{Powell2002, Mammoto2013, Shiwarski2020, Ingber2006, Donnelly2010, Rangarajan2014}. Thus, it is imperative to model the adaptation of muscle force output from biohybrid systems during the design process if they are to be functionally controlled to accomplish operational tasks. The current state-of-the-art for modeling biohybrid robots is predominantly focused on modeling their mechanics. Finite element analysis (FEA) tools like ANSYS have been used to capture the mechanics of biohybrid systems \cite{Cvetkovic2014, Webster2016, Schaffer2022AConstructs}. However, FEA can fall short as a model and design tool for the large deformations of soft systems \cite{Zhang2019ModelingArchitectures}, often carrying a large computational cost for model convergence. Similarly, analytical modeling has been used to describe biohybrid systems \cite{Webster2016, raman2016optogenetic}. However, for increasingly sophisticated systems, analytical modeling can be insufficient to capture the geometric and mechanical complexity of such systems, and analytical models must be reformulated for new designs. In addition to modeling, controlling biohybrid systems also poses a challenge to researchers. 
Generally, biohybrid robotic control has been bespoke. It often makes use of the time-tested technique known as ``graduate student with tweezers" \cite{sarah_ted}. More methodical, bioinspired approaches show promise as control schemes for biohybrid robots, including synthetic nervous systems \cite{RaveshSNS, vickieSNS} as does reinforcement learning \cite{Naughton2020Elastica:Control, ppo_paper}, especially for high degree of freedom systems with distributed actuation experiencing large deformations.

Here, we demonstrate a preliminary step toward modeling biohybrid embodied intelligence through the unification of mechanics, control, and, for the first time, biohybrid adaptability from exercise history. Using our previously established architecture of a distributed worm-inspired lattice as a testbed for our investigation \cite{schaffer2023tall}, we made several key contributions. Firstly, we incorporate biomimetic strain- and contraction force-dependent muscle actuator adaptability in the modeling of biohybrid robots. Secondly, we implement an off-the-shelf reinforcement algorithm (PPO) to successfully coordinate the independent contraction of 42 muscles distributed on a lattice worm structure to steer it towards 8 distinct targets while incorporating adaptability. We report that the adaptive case outperforms the non-adaptive case, both in terms of training time and maximum rewards By capturing the use-based adaptation of muscle actuators paired with a controller capable of coordinating their contractions, biohybrid robotics is moved one step closer to operational performance.

\section{Methods}

All simulations were conducted using PyElastica \cite{Zhang2019ModelingArchitectures,Gazzola2018ForwardFilaments}. PyElastica is an open-source software package for simulating assemblies of soft, slender bodies by making use of the Cosserat Rod theory \cite{cosserat1909theorie}. Briefly, deformable bodies are modeled as 1D Cosserat rods. This captures their 3D dynamics, accounting for all modes of deformation (\textit{i.e.} bending, twisting, stretching, and shearing) while remaining computationally lightweight. One key benefit of the Cosserat representation is that its complexity scales linearly with axial resolution, while finite element methods scale cubically \cite{Gazzola2018ForwardFilaments}. Scaling linearly has a benefit in massively reducing computational cost, especially as system complexity increases. For a detailed explanation of Cosserat Rod Theory, readers are encouraged to visit \url{cosseratrods.org}, a site managed by the maintainers of PyElastica.

\begin{figure}[t!]
    \centering
    \includegraphics[width=1.0\textwidth]{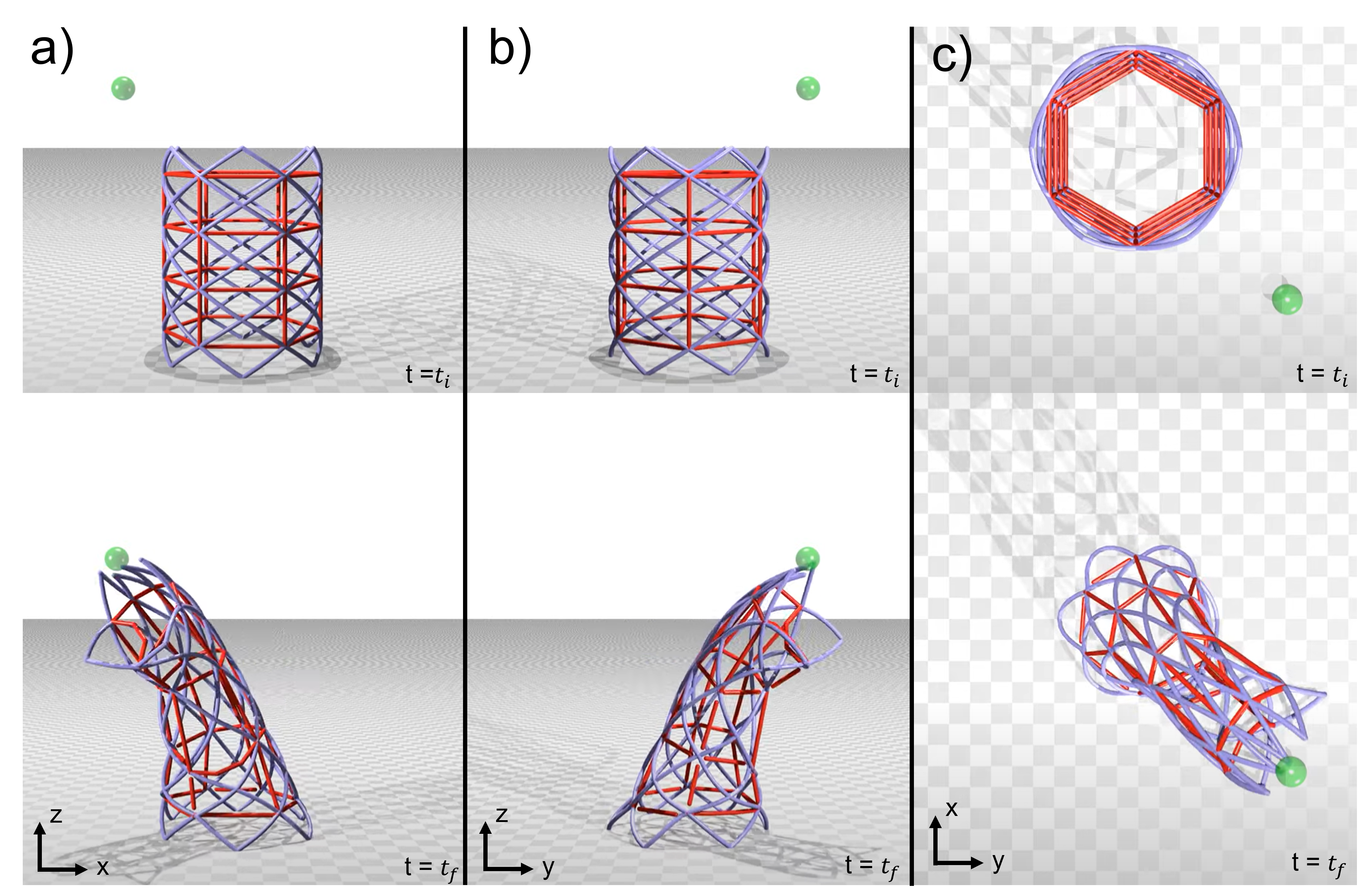}
    \caption[RL agent coordinates 42 muscles and achieves successful lattice worm reaching]{Reinforcement learning control policy coordinates 42 distributed muscles and achieves successful lattice worm reaching. A lattice worm, shown from three different views, starts in an undeformed state in the top panels and is deformed by muscle actuation in the bottom panels. \textbf{a)} Front view of lattice worm in \textit{x-z} plane. \textbf{b)} Side view of lattice worm in \textit{y-z} plane. \textbf{c)} Top view of lattice worm in \textit{y-x} plane. For each view, the top panel shows the state of the lattice worm in the initial state at $t=t_{0}$, and the bottom panel shows the lattice worm in the final state at $t=t_{f}$. }
    \label{fig:reaching_fig}
\end{figure}

\begin{figure}[t!]
    \centering
    {\includegraphics[width=0.9\textwidth]{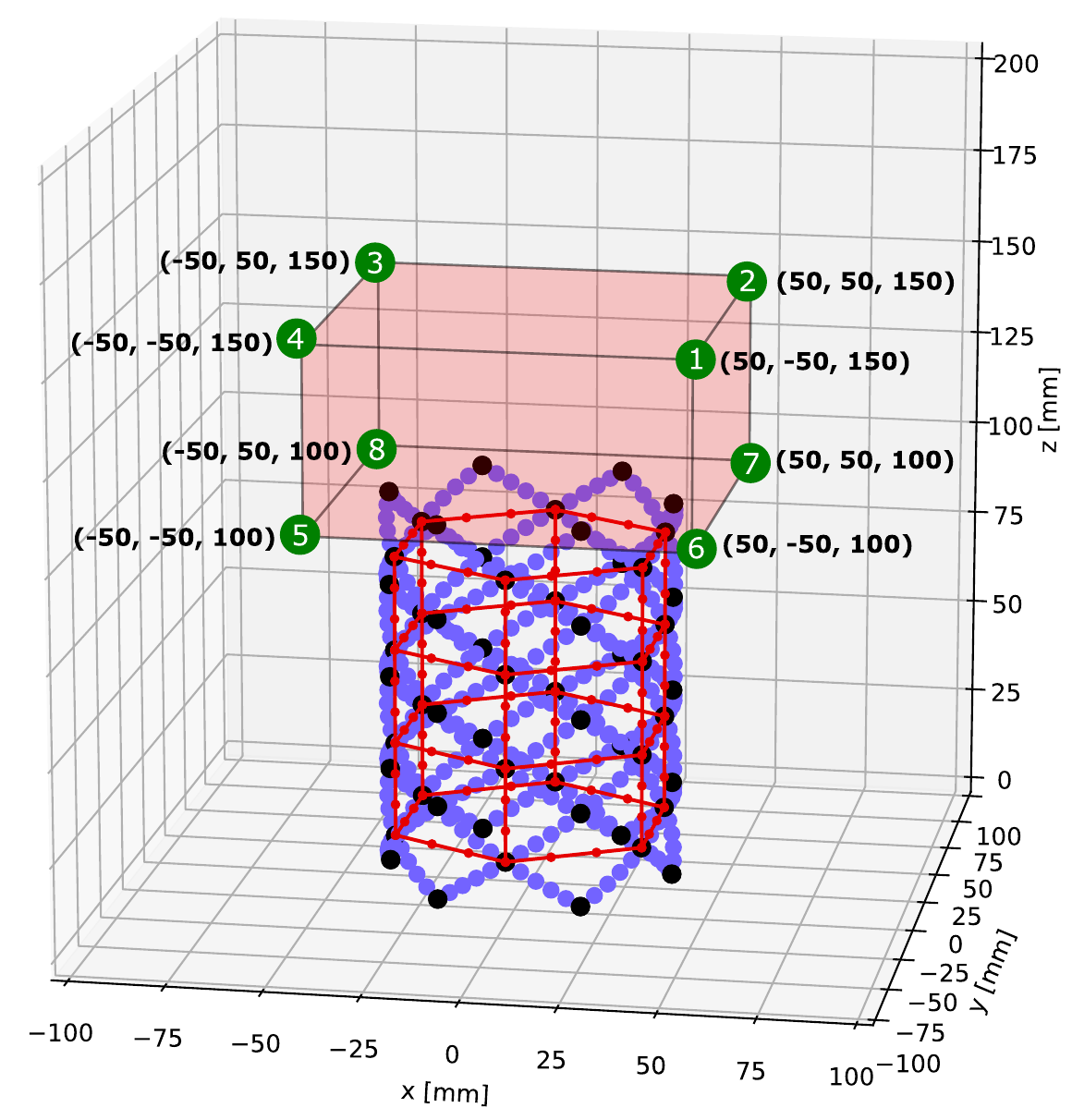}}
    \caption[Lattice worm robot and target locations]{Lattice worm robot and target locations. The robot is discretized, with each element represented as a sphere. Purple represents the structural rods. Red represents the muscle rods. Black represents the connection points between rods. Green numbered dots are the locations of eight target locations.}
    \label{fig:prism}
\end{figure}

The lattice worm architecture used for this study is an extension of work detailed in \cite{schaffer2023tall}. In that previous work, we detailed a parametrically defined lattice worm patterned with between four and six muscles hand-placed that could achieve bulk extension, compression, and bending of its body structure. Building off of that foundation, in this work, we patterned a total of 42 muscles and were able to achieve targeted reaching to arbitrary locations through a learning-based controller to coordinate muscle actuator activation. When any of these muscle actuators contract, they produce a tension that deforms the lattice locally. By coordinating these local contractions, desirable behaviors of the lattice worm are achievable. The desirable behavior in this study is maneuvering to a specified target location in 3D space. Eight corners of a rectangular prism were selected as the target locations, illustrated in Figure \ref{fig:prism}. Figure \ref{fig:reaching_fig} illustrates the coordinated deformation of the lattice worm when reaching for a target.

\subsection{Reinforcement Learning Model Setup}
Reinforcement learning is a machine learning technique in which an agent iteratively learns to select the best actions that maximize a reward after observing an environment \cite{kim2021review}. In our case, the agent is the controller for the robot, the actions are muscle actuator activation strength, the environment is the PyElastica simulation environment, and the maximum reward is achieved by touching the green target sphere. Through many iterations and by taking progressively less random actions, an agent is able to learn which actions it should take in a given state to maximize a reward. In our case, this manifests as an agent first guessing random muscle activations and then learning how the state evolves as a result of those actions, progressively updating its control policy. Through this iterative approach, our agents discover which muscle activation intensities steer the lattice worm closest to the target. A natural byproduct of the agent learning a control policy for a performant behavior such as targeted reaching is that the agent also reveals the relative importance of muscles to the task based on their activation intensity (\textit{i.e.}, muscles that are not activated are not needed for the task and vice versa).

To coordinate the contractions of the 42 muscles and identify their relative importance, we deployed proximal policy optimization (PPO) \cite{ppo_paper}. PPO is a model-free policy gradient method and is considered a state-of-the-art RL algorithm for continuous control with demonstrated performance in a variety of tasks \cite{truly_ppo, Naughton2020Elastica:Control}. To ensure proper implementation of the algorithms, we used the Stable Baselines3 \cite{stable-baselines3} implementation of PPO paired with a custom environment compatible with the Gymnasium API \cite{towers_gymnasium_2023} (formerly OpenAI Gym\cite{brockman2016openai}). We used the default model parameters \cite{stable_baselines3_ppo} with the exception of seeds and $n\_steps=2$, where $n\_steps$ \cite{stable_baselines3_ppo} is the number of steps to run for each environment per update.

The reward function was adapted from \cite{Naughton2020Elastica:Control} and penalizes the distance between the terminus of the robot to the green target sphere (Figure \ref{fig:prism}): 

\[ 
R = -n^2 + \phi(n) 
\]

\noindent where $n$ is the 3D Euclidean distance between the terminus of the lattice worm robot given by \(n = \lVert x_n - x_t \rVert \), and $\phi$ is two-tiered bonus reward:

\[
\phi(n) = 
 \begin{cases} 
 0.5 & d<n<2d 
 \\ 2.0 & n<d 
 \end{cases}
\]

\noindent where $d$ = 1 mm. Additionally, because the allowed action space is capable of causing the simulation to become unstable, the simulation reward was set to -2 any time a NaN was detected in the state information, an indication of episode instability. Detection of a NaN would cause the episode to end. The state, which is the view of the environment that the agent sees, was set as \( S = [x_{a}, v_{a}, A, \lambda, x_{t}]\), where \(x_{a}, v_{a}\) describe the three-dimensional position and velocity of the lattice worm, $A$ is the previous action set, and $\lambda$ is the adaptation-dependent force ceiling for each muscle. When adaptation is enabled, this maximum force can increase between episodes as detailed in Section \ref{sec:Adaptive Muscle Force Output}.

The action space for the agent is to select the activation $A_m \in [0, 1]$ for each of the 42 muscles. It is important to note that the agent does not select the force the muscle will produce in a given episode. Instead, it selects an activation of the muscle, which is then multiplied by the maximum force output for that muscle, depending on its use history. That is, the force produced by a muscle in a given episode is $F_{m, i} = A_{m,i} \cdot \lambda_{m,i}$, where $F_{m,i}$, $A_{m,i}$, $\lambda_{m,i}$ are the force output, muscle activation, and force ceiling for a given muscle during a given episode, respectively. To consistently produce the same amount of force for a given adaptive muscle, the agent must learn to modulate its activation $A$ as $\lambda$ changes between episodes.

We did not perform hyperparameter tuning, which suggests that the rewards reported might not be the maximum attainable for each adaptation status and target combination. However, the utility of reinforcement learning in this work is not to elucidate which hyperparameters are best for these particular cases but rather to demonstrate reinforcement learning algorithms as potentially helpful design partners in creating and modeling operational distributed adaptive biohybrid robots.

\subsection{PyElastica Simulation Environment}

Critical parameters that define the simulation environment are detailed in Table \ref{tab:sim_param_table}. These parameters define the geometry, discretization, material properties, and adaptation coefficients.

\subsubsection{Adaptive Muscle Force Output}
\label{sec:Adaptive Muscle Force Output}

To better capture the adaptation of muscle actuators in response to use history, we developed an adaptation function that updates the force ceiling $\lambda$ of each muscle proportional to its use history. This force ceiling is the maximum force a muscle can produce if the reinforcement learning agent selects an activation value of 1 for a given episode. $\lambda$ was calculated as,

\[
 \begin{aligned}
    \lambda_{m, i} &= \min \left( \alpha_{m, i} \cdot \lambda_{m, i-1}, \ \ 2\lambda_{0} \right) \\
    \alpha_{m, i} &=  1 +  \beta  \left| \epsilon_{m, i-1} \right| + \gamma \left| F_{m, i-1} \right|
\end{aligned}
\]

\noindent where $\alpha_{i}$ is the adaptation coefficient, $\beta$ is the strain coefficient,  $F_{i-1}$ and $\epsilon_{max}$ are the maximum stress experienced and the force produced by the muscle in the previous episode, respectively. $\lambda_{0}$ is the initial force ceiling at $i=0$. The subscripts $i$  and $m$ refer to the particular episode and muscle, respectively. These equations formalize incrementing the force ceiling of a muscle proportional to the strain it experienced and the force it produced in the previous episode, up to twice its starting value at the first episode.

\begin{table}[t!]
\centering
\caption[Updated Lattice Worm Parameters]{Parameters that define the lattice worm morphology, material model, adaptability, and simulation variables. Similar parameters were used in \cite{schaffer2023tall}.}
\label{tab:sim_param_table}

\begin{tabular}{l|rl}
\textbf{Simulation Parameters} & \textbf{Value} & \textbf{Units} \\ \hline
Structural rod elements        & 40            & elements      \\
Muscle rod elements            & 2             & elements      \\
Lattice height ($h$)             & 100           & mm            \\
Lattice diameter ($d$)           & 75            & mm            \\
Damping coefficient            & 35            & mNs/m         \\
Muscle Young's modulus         & 25            & kPa           \\
Structure Young's modulus      & 70            & kPa           \\
Muscle Poisson's ratio         & 0.5           &               \\
Structure Poisson's ratio      & 0.5           &               \\
Structure density              & 1070          & kg/m$^3$      \\
Muscle density                 & 1060          & kg/m$^3$      \\
Structure rod radius           & 10             & mm            \\
Muscle rod radius              & 5             & mm            \\
Connection stiffness ($k$)       & 100           & mN/mm         \\
Initial force ($\lambda_{0}$)    & 2000          &              \\
Strain coefficient ($\beta$)     & $10^{-6}$     &               \\
Previous force coefficient ($\gamma$)    & $4\times10^{-8}$ &             \\
\textit{n\_steps}                          & 2

\end{tabular}
\end{table}

\subsection{Experiments}

To elucidate the effect of muscle adaptability on lattice worm performance, we trained a total of 80 reinforcement learning agents. Each agent was learning in either an environment that featured muscle adaptation or no adaptation. Furthermore, each agent was assigned to navigate the lattice worm towards one of eight target locations illustrated in Figure \ref{fig:prism}. For each adaptation/target location pairing, 5 agents were trained, each with different seeds initializing their deep policy neural network \cite{ppo_paper}.

Computational experiments were prototyped on a local workstation running Ubuntu 22.04, equipped with an AMD Ryzen 7 3700 8-core/16-thread processor and NVIDIA GeForce RTX 2080 SUPER GPU. The reported experiments were run on the GPU-Shared partitions of the Bridges-2 \cite{bridges2} cluster at the Pittsburgh Supercomputing Center. This features NVIDIA Tesla V100-32GB SXM2 and 2 Intel Xeon Gold 6248 ``Cascade Lake" CPUs, 20 cores per CPU, 2.50-3.90 GHz. The hardware used from computational prototyping is mentioned as all experiments conducted are feasible to run on the workstation, with each individual agent training taking a similar amount of time. 

\section{Results and Discussion}

To evaluate the reaching performance for lattice worms featuring muscle adaptability and lattice worms without adaptability, we trained 5 differently seeded agents for each corner/adaptation combination for a total of 80 agents. Their performance is detailed in Figure \ref{fig:rewards-plot}. We find that in all cases, the agent is able to learn a successful control policy to steer the lattice worm towards the target location. Each training run was given 48 hours to complete 8,500 episodes, though most training runs were completed in under 45 hours. This is a relatively small number of episodes compared to the several million episodes common in the reinforcement learning literature. This suggests that better performance is likely to be achieved if trained on more episodes \cite{Naughton2020Elastica:Control}. However, achieving globally optimal performance on these particular tasks is outside the scope of this work and, as such, was not pursued. Instead, our goal was to implement adaptability into biohybrid robot design, modeling, and control.

\subsection{Comparing Adaptable and Non-adaptable Agent Performance }

\begin{figure}[t]
    \centering
    \includegraphics[width=1.0\linewidth]{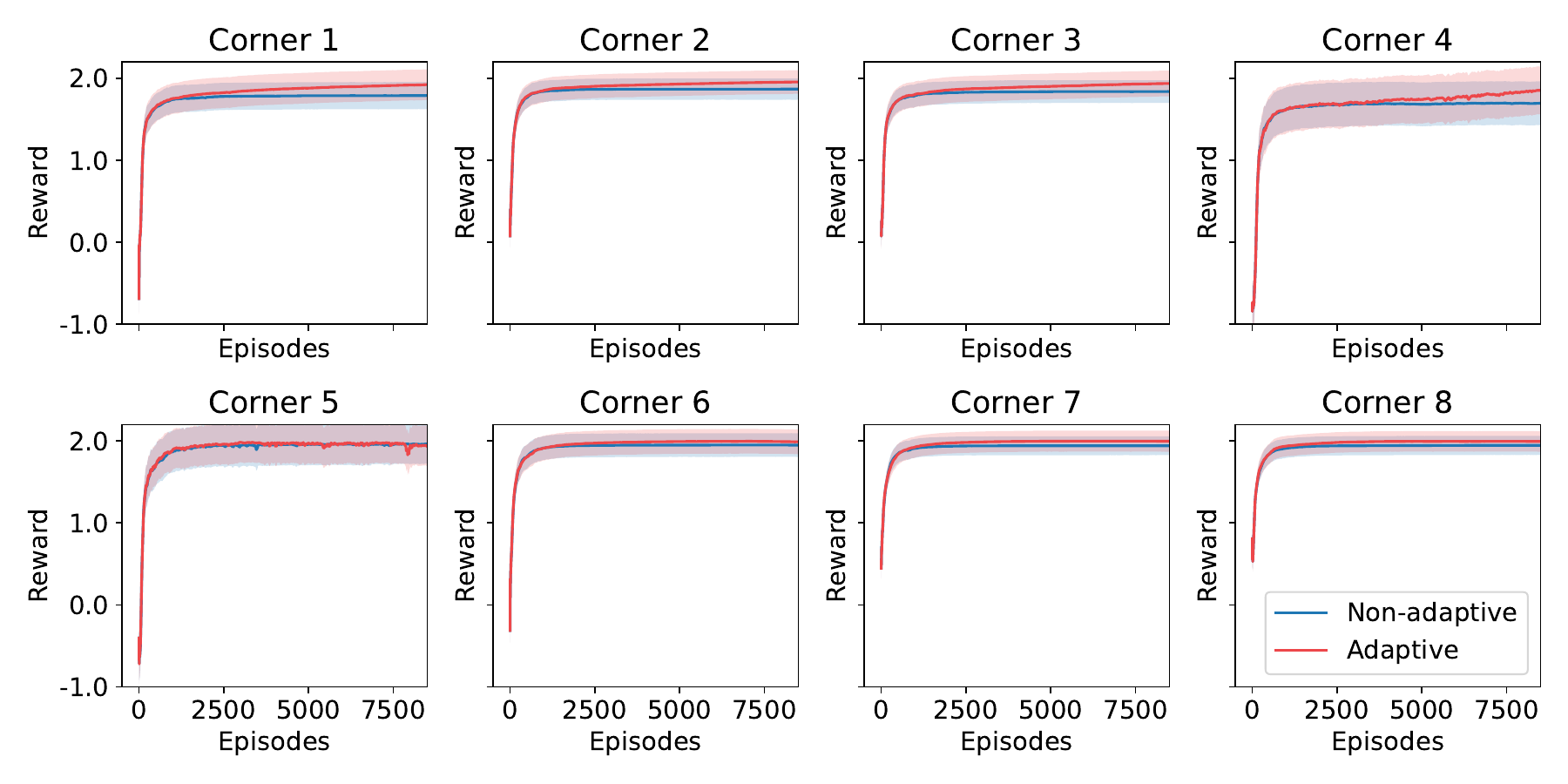}
    \caption[Training results for lattice worm reaching]{Training results for lattice worm reaching for eight distinct corners. Each plot features the training results for both the adaptive and non-adaptive muscle cases for 8,500 episodes. Curves are the rolling 50 sample average of the combined results across 5 seeds, with the shaded regions representing one standard deviation. Episodes resulting in simulation instability were excluded. The legend in Corner 8 is for all corners.}
    \label{fig:rewards-plot}
\end{figure}

Intuitively, we expected to find that the adaptive case would be slower (\textit{i.e.}, take more episodes) to learn given that the agent must not only contend with coordinating the actuation of 42 independent muscles, but also each of those muscle force ceilings is potentially changing every episode. However, our results in Figure \ref{fig:rewards-plot} suggest this is not the case. Early in the training, the performance between the adaptive and non-adaptive agents is quite similar as all agents have similar force ceilings for their muscles in early episodes. However, as the adaptable agent exercises muscles that are important to the task, those muscles become stronger, and the agent is able to maneuver the lattice worm to the goal more successfully. We believe this result to be the combination of two phenomena: (1) the remarkable ability of robust reinforcement learning algorithms such as PPO to ingest complex state spaces and select optimal actions, (2) adaptation may act as an ersatz implementation of curriculum learning \cite{curriculum-learning}.

In curriculum learning, the environment presented to the agent starts off easier and gets progressively more challenging as episodes elapse. By starting off easy, the agent can learn a policy more readily. As the environment becomes more challenging, the agent already has the solution to a similar but easier problem. In this way, curriculum learning aims to help reinforcement learning agents learn better. Our adaptability implementation serves as an ersatz curriculum learning strategy. Trying to coordinate the contraction of muscles with higher force ceilings is more challenging because contracting these muscles more readily leads to simulation instabilities, which are harder for the agent to learn from. By gradually increasing the maximal force output of the muscles across episodes, the agent is effectively exposed to a curriculum.

\begin{figure}[t!]
    \centering
    \includegraphics[width=1.0\linewidth]{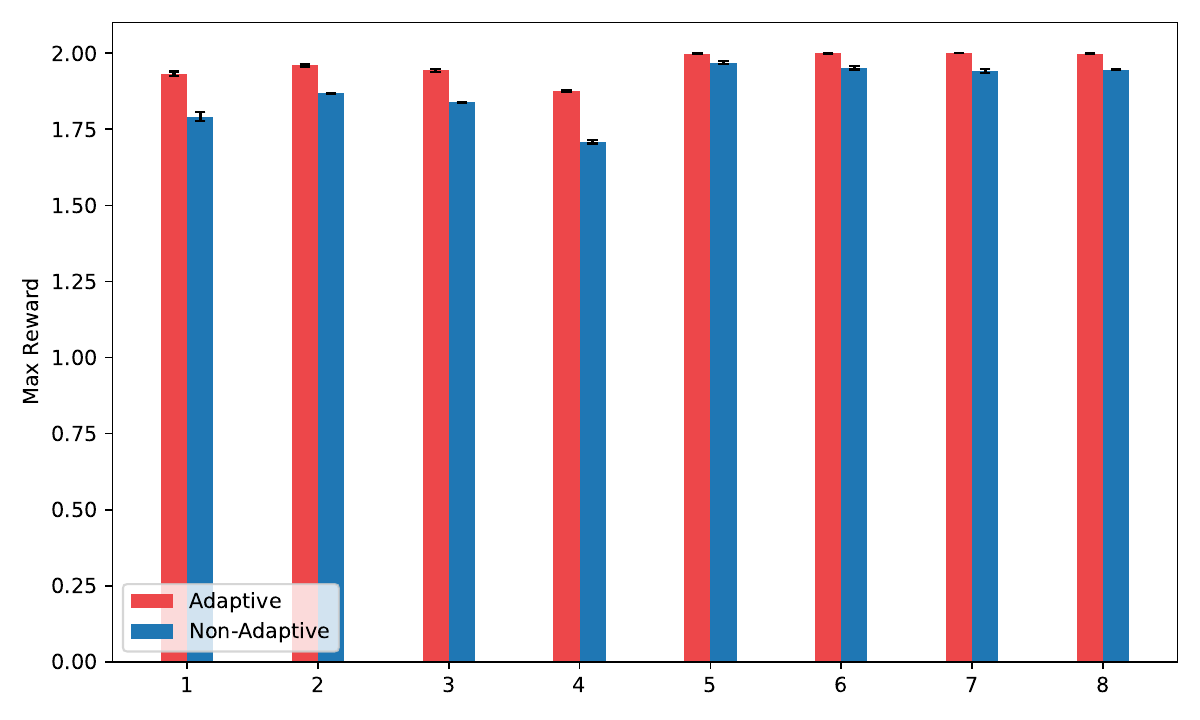}
    \caption[Maximum reward for each corner and environment type]{Reward maximum averaged across seeds for each of the eight target corners. Bars are the mean, and whiskers are one standard deviation.}
    \label{fig:max-averg-rewards-plot}
\end{figure}

We also expected to see that the maximal reward achieved by adaptable agents is greater because adaptable agents can muster greater muscle actuator forces in their later training episodes. This is reflected in our results, as shown in Figure \ref{fig:max-averg-rewards-plot}, illustrating that for all corners, the adaptable agents can achieve higher maximum rewards than their non-adaptable counterparts. This difference is more pronounced for targets that are harder to reach because they are farther away, \textit{i.e.} corners 1, 2, 3, and 4 (Figure \ref{fig:prism}).

\subsection{Adaptation Across Episodes}

\begin{figure}[t!]
    \centering
    \adjustbox{trim=0 0 0 {.05\height},clip}
    {\includegraphics[width=0.99\textwidth]{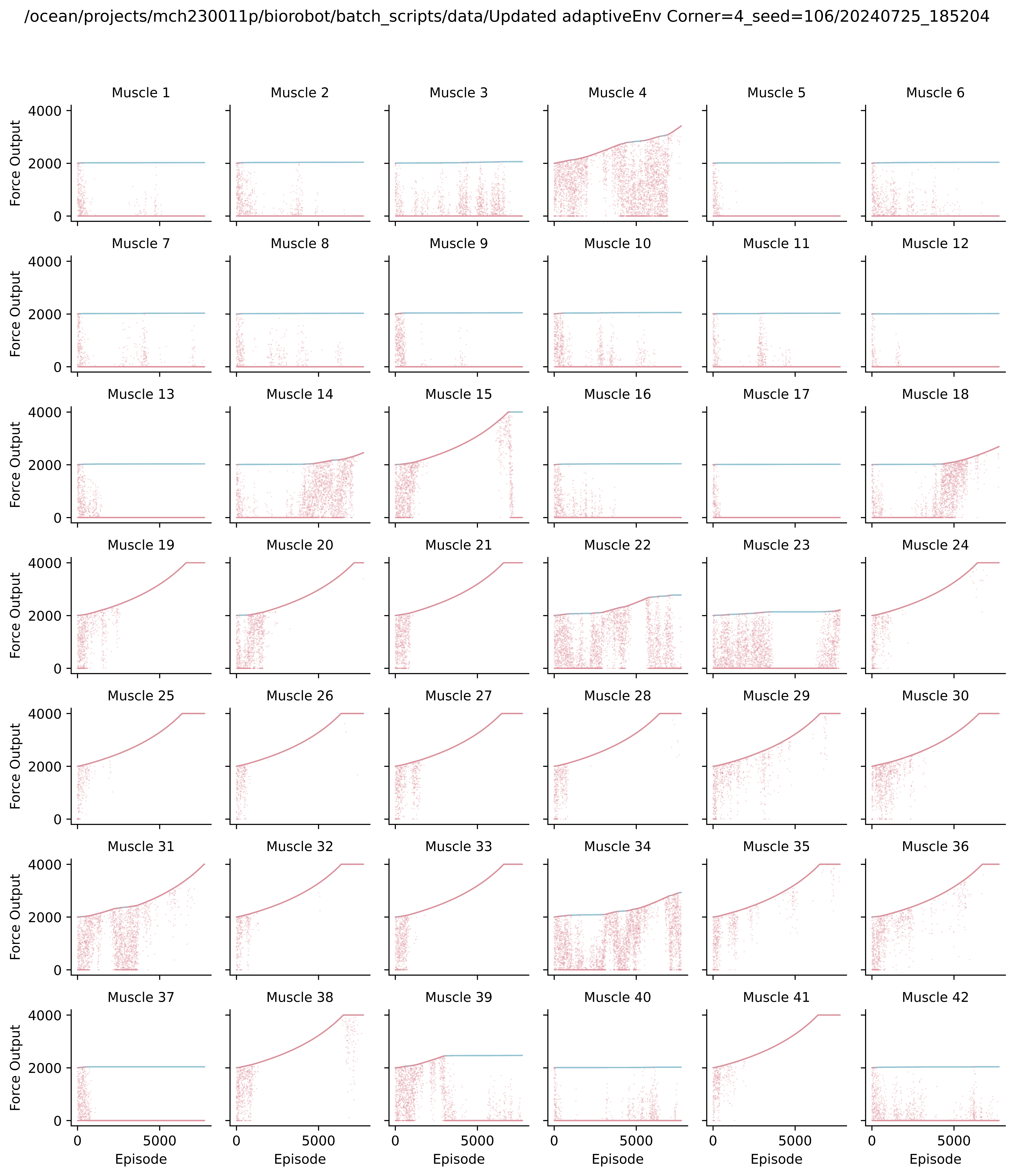}}
    \caption[Representative muscle adaptation across learning episodes]{Representative muscle adaptation across learning episodes. For each of the 42 muscles patterned on the lattice worm, the blue points are dense enough to form a line and represent the force ceiling for that muscle in a given episode. The red points represent the force a given muscle produced during a given episode. Force increases result from exercise in the previous episodes quantified through strain and use of the muscle.  For muscles that are used in previous episodes, the force ceiling increases. Data from adaptive lattice worm reaching for corner 4, seed=106.}
    \label{fig:adaptation_plot}
\end{figure}

From our adaptation scheme, we expect to see that muscles that are used often become stronger. 
Figure \ref{fig:adaptation_plot} details a representative history of muscle adaptation across the episodes of an adaptive agent learning a control policy. Indeed, we see that for muscles that are used in previous episodes, the force ceiling increases. If a muscle stops being used in subsequent episodes, its force ceiling plateaus, as is the case for Muscle 39. Additionally, if a muscle is used more often across many episodes, its force ceiling is capped at twice the initial value, like for Muscle 27. These features of muscle adaption are intentionally codified in our adaptation equations and their emergence in our results supports that we successfully implemented our desired bioinspired muscle adaptation scheme.

\subsection{Identifying Required Muscles From Activation Intensity}
In addition to coordinating the contraction of muscles, leveraging reinforcement learning helps identify which muscles are being used to complete a given task. In the targeted reaching task illustrated in Figure \ref{fig:reaching_fig}, there are 42 possible muscles the agent could have selected to contract. However, not all muscles are strictly needed to complete a given task. Identifying which muscles are needed is a combinatorically-hard problem given by \( \binom{42}{k} \), where \( k \) is the number of muscles to be fabricated. To illustrate how large this space is by way of example, when \( k = 20 \), the number of combinations is greater than 513 billion. 

\begin{figure}[h!]
    \centering
    \includegraphics[width=0.6\textwidth]{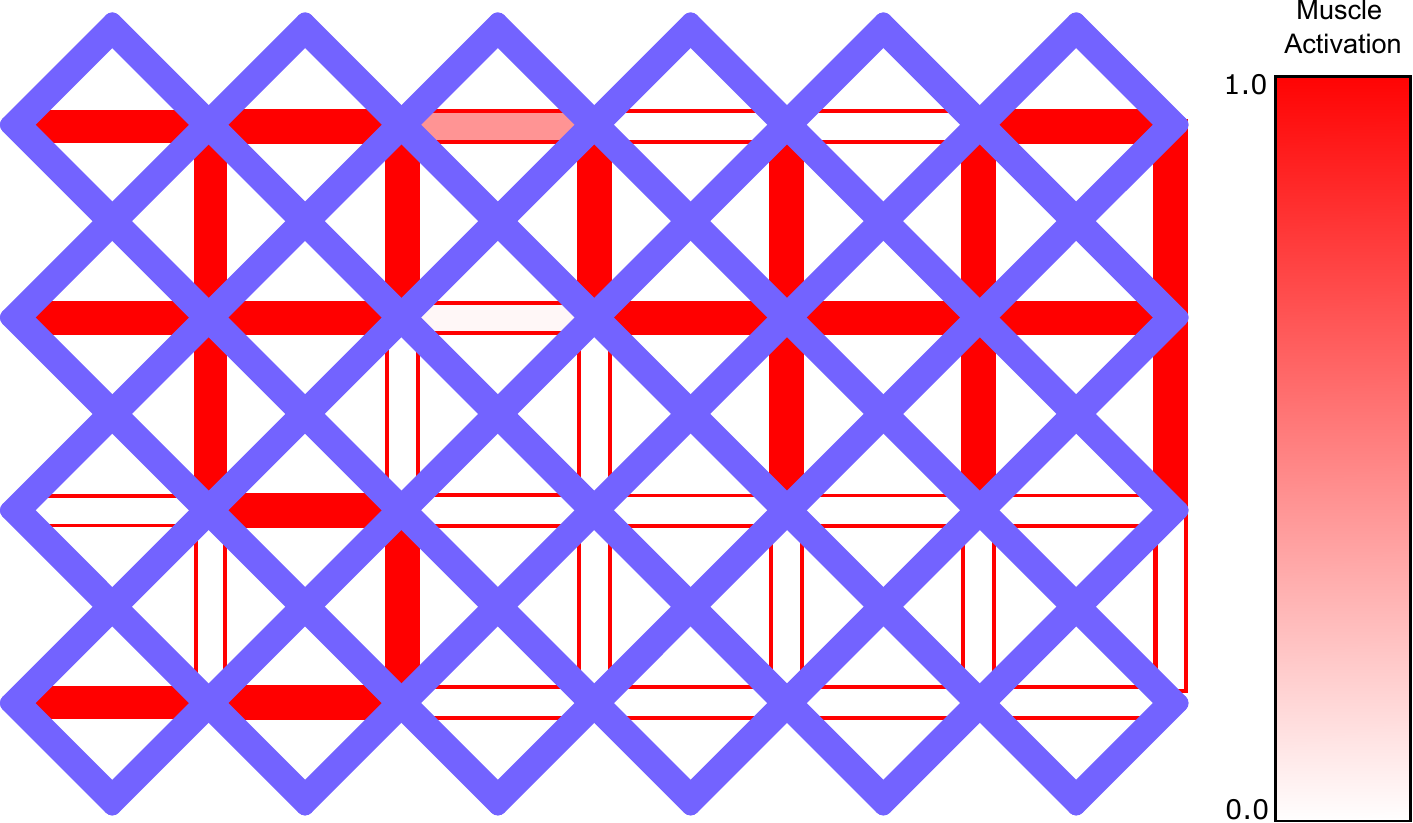}
    \caption[Muscle activation heatmap]{Muscle activation heatmap. The lattice worm is unwrapped to form a 2D lattice of structure rods (\textit{purple}) and muscle rods (\textit{red}). The redness of the muscle corresponds to its normalized activation intensity.}
    \label{fig:unwrapped-lattice}
\end{figure}

By learning a successful control policy for the task, the reinforcement learning agent also elucidates which muscles were required to complete that task based on which muscles were activated. Figure \ref{fig:unwrapped-lattice} shows an unwrapped 2D lattice representation of the 3D lattice worm reaching task in Figure \ref{fig:reaching_fig}. This procedurally generated figure elucidates which muscles were needed for task completion and thus can allow researchers to identify in simulation which muscles are needed before the arduous and expensive fabrication of physical devices through trial and error.

\subsection{Current Adaptation Implementation Limitations}

This work captures a bioinspired strain- and force-dependent adaptability based on past usage in the modeling of a muscle actuator's force output. In our implementation (Section \ref{sec:Adaptive Muscle Force Output}), the hyperparameters \(\beta\) and \(\gamma\) define the effect of the strain experienced and forces produced by a muscle in the previous episode on incrementing the force ceiling of that same muscle in the current episode. These adaptability hyperparameters are inspired by experimental adaptation studies in the literature \cite{Powell2002, Mammoto2013, Shiwarski2020, Ingber2006, Donnelly2010, Rangarajan2014}. However, the particular values chosen for \(\beta\) and \(\gamma\) are not mapped directly from a specific experimental study. Future experimental work that elucidates the relationship between the myriad factors (\textit{e.g.} cell source, extracellular matrix composition, media composition, etc.) and muscle force output should be used to improve the fidelity of our muscle adaptation model and, by extension, our biohybrid robot modeling tool as a whole. Furthermore, this work does not capture the degradation phenomena associated with muscle actuation, namely intra-episode muscle fatigue from intense use or inter-episode muscle atrophy from disuse. These short- and long-term phenomena, respectively, will be important to capture in models that aim to elucidate the longitudinal performance of biohybrid robots.

\section{Conclusion}

Realizing the full potential of biohybrid robotics requires modeling, designing, and controlling their adaptive actuation within distributed contexts. In this work, we demonstrate a preliminary step toward RL-based co-design and modeling of biohybrid embodied intelligence through the unification of mechanics, control, and, for the first time, biohybrid adaptability from exercise history. Updating the architecture of a distributed worm-inspired lattice from our previous work as a testbed for our current investigation, we made several key contributions. Firstly, we incorporate biomimetic strain- and force-dependent adaptability based on past usage in the modeling of muscle actuator's force output. Additionally, we utilized an existing robust reinforcement algorithm as both a controller and co-design tool. As a controller, it coordinates the independent contraction of 42 muscles distributed on a lattice worm structure to successfully steer it towards eight distinct targets while incorporating this adaptability.  As a co-design tool, it enables users to identify which muscles are important to accomplishing a given task, allowing a fabricator to focus their resources on making the actuators that underpin performance. Our results show that adaptive agents outperform non-adaptive agents in terms of maximum rewards and training time. By capturing the use-based adaptation of muscle actuators paired with a controller capable of coordinating their contractions, biohybrid robotics is moved one step closer to operational performance.

\section{Acknowledgements}

This work was supported by the NSF GRFP under grant No. DGE1745016 and by the NSF Faculty Early Career Development Program (ECCS-2044785). This work used Bridges-2 at Pittsburgh Supercomputing Center through allocation MCH230026 from the Advanced Cyberinfrastructure Coordination Ecosystem: Services \& Support (ACCESS) program, which is supported by National Science Foundation grants \#2138259, \#2138286, \#2138307, \#2137603, and \#2138296. Also thank you to T.J. Olesky at the Pittsburgh Supercomputing Center (PSC) for his help spinning up  S.S. in HPC and raising the priority of S.S.'s batch jobs when requested.


\bibliographystyle{IEEEtran}
\bibliography{all_bib}

\begin{thebibliography}{10}
\providecommand{\url}[1]{#1}
\csname url@samestyle\endcsname
\providecommand{\newblock}{\relax}
\providecommand{\bibinfo}[2]{#2}
\providecommand{\BIBentrySTDinterwordspacing}{\spaceskip=0pt\relax}
\providecommand{\BIBentryALTinterwordstretchfactor}{4}
\providecommand{\BIBentryALTinterwordspacing}{\spaceskip=\fontdimen2\font plus
\BIBentryALTinterwordstretchfactor\fontdimen3\font minus \fontdimen4\font\relax}
\providecommand{\BIBforeignlanguage}[2]{{%
\expandafter\ifx\csname l@#1\endcsname\relax
\typeout{** WARNING: IEEEtran.bst: No hyphenation pattern has been}%
\typeout{** loaded for the language `#1'. Using the pattern for}%
\typeout{** the default language instead.}%
\else
\language=\csname l@#1\endcsname
\fi
#2}}
\providecommand{\BIBdecl}{\relax}
\BIBdecl

\bibitem{raman2024biofabrication}
R.~Raman, ``Biofabrication of living actuators,'' \emph{Annual Review of Biomedical Engineering}, vol.~26, no.~1, pp. 223--245, 2024.

\bibitem{raman2024soft}
R.~Raman and C.~Laschi, ``Soft robotics for human health,'' \emph{Device}, vol.~2, no.~7, 2024.

\bibitem{Raman2017}
R.~Raman, L.~Grant, Y.~Seo, C.~Cvetkovic, M.~Gapinske, A.~Palasz, H.~Dabbous, H.~Kong, P.~P. Pinera, and R.~Bashir, ``{Damage, Healing, and Remodeling in Optogenetic Skeletal Muscle Bioactuators},'' \emph{Advanced Healthcare Materials}, vol.~6, no.~12, pp. 1--9, 2017.

\bibitem{Ricotti2017BiohybridCells}
L.~Ricotti, B.~Trimmer, A.~W. Feinberg, R.~Raman, K.~K. Parker, R.~Bashir, M.~Sitti, S.~Martel, P.~Dario, and A.~Menciassi, ``{Biohybrid actuators for robotics: A review of devices actuated by living cells},'' \emph{Science Robotics}, vol.~2, no.~12, 11 2017.

\bibitem{Webster-Wood2017}
V.~A. Webster-Wood, O.~Akkus, U.~A. Gurkan, H.~J. Chiel, and R.~D. Quinn, ``{Organismal engineering: Toward a robotic taxonomic key for devices using organic materials},'' \emph{Science Robotics}, vol.~2, no.~12, pp. 1--19, 2017.

\bibitem{morimoto2020biohybrid}
Y.~Morimoto and S.~Takeuchi, ``Biohybrid robot powered by muscle tissues,'' \emph{Mechanically Responsive Materials for Soft Robotics}, pp. 395--416, 2020.

\bibitem{Raman2017b}
R.~Raman, C.~Cvetkovic, and R.~Bashir, ``{A modular approach to the design, fabrication, and characterization of muscle-powered biological machines},'' \emph{Nature Protocols}, vol.~12, no.~3, pp. 519--533, 2017.

\bibitem{Powell2002}
C.~A. Powell, B.~L. Smiley, J.~Mills, and H.~H. Vandenburgh, ``{Mechanical stimulation improves tissue-engineered human skeletal muscle},'' \emph{American Journal of Physiology - Cell Physiology}, vol. 283, no. 5 52-5, pp. 1557--1565, 2002.

\bibitem{Mammoto2013}
T.~Mammoto, A.~Mammoto, and D.~E. Ingber, ``{Mechanobiology and Developmental Control},'' \emph{Annual Review of Cell and Developmental Biology}, vol.~29, no.~1, pp. 27--61, 2013.

\bibitem{Shiwarski2020}
D.~J. Shiwarski, J.~W. Tashman, A.~F. Eaton, G.~Apodaca, and A.~W. Feinberg, ``{3D printed biaxial stretcher compatible with live fluorescence microscopy},'' \emph{HardwareX}, vol.~7, p. e00095, 2020.

\bibitem{Ingber2006}
D.~E. Ingber, ``{Cellular mechanotransduction: putting all the pieces together again},'' \emph{FASEB Journal}, vol.~20, no.~8, pp. 811--827, 2006.

\bibitem{Donnelly2010}
K.~Donnelly, A.~Khodabukus, A.~Philp, L.~Deldicque, R.~G. Dennis, and K.~Baar, ``{A Novel bioreactor for stimulating skeletal muscle in vitro},'' \emph{Tissue Engineering - Part C: Methods}, vol.~16, no.~4, pp. 711--718, 2010.

\bibitem{Rangarajan2014}
S.~Rangarajan, L.~Madden, and N.~Bursac, ``{Use of Flow, Electrical, and Mechanical Stimulation to Promote Engineering of Striated Muscles},'' \emph{Annals of Biomedical Engineering}, vol.~42, no.~7, pp. 1391--1405, 2014.

\bibitem{Cvetkovic2014}
C.~Cvetkovic, R.~Raman, V.~Chan, B.~J. Williams, M.~Tolish, P.~Bajaj, M.~{Selman Sakar}, H.~H. Asada, M.~Taher, A.~Saif, and R.~Bashir, ``{Three-dimensionally printed biological machines powered by skeletal muscle},'' \emph{Proceedings of the National Academy of Sciences}, vol.~11, no.~28, pp. 10\,125--10\,130, 2014.

\bibitem{Webster2016}
V.~A. Webster, S.~G. Nieto, A.~Grosberg, O.~Akkus, H.~J. Chiel, and R.~D. Quinn, ``{Simulating muscular thin films using thermal contraction capabilities in finite element analysis tools},'' \emph{Journal of the Mechanical Behavior of Biomedical Materials}, vol.~63, pp. 326--336, oct 2016.

\bibitem{Schaffer2022AConstructs}
S.~Schaffer, J.~S. Lee, L.~Beni, and V.~A. Webster-Wood, ``{A Computational Approach for Contactless Muscle Force and Strain Estimations in Distributed Actuation Biohybrid Mesh Constructs},'' \emph{Biomimetic and Biohybrid Systems}, vol. 13548, pp. 140--151, 2022.

\bibitem{Zhang2019ModelingArchitectures}
X.~Zhang, F.~K. Chan, T.~Parthasarathy, and M.~Gazzola, ``{Modeling and simulation of complex dynamic musculoskeletal architectures},'' \emph{Nature Communications}, vol.~10, no.~1, 12 2019.

\bibitem{raman2016optogenetic}
R.~Raman, C.~Cvetkovic, S.~G. Uzel, R.~J. Platt, P.~Sengupta, R.~D. Kamm, and R.~Bashir, ``Optogenetic skeletal muscle-powered adaptive biological machines,'' \emph{Proceedings of the National Academy of Sciences}, vol. 113, no.~13, pp. 3497--3502, 2016.

\bibitem{sarah_ted}
\BIBentryALTinterwordspacing
S.~Bergbreiter, ``{Why I make robots the size of a grain of rice},'' Sep 2018. [Online]. Available: \url{https://youtube.com/watch?v=PtBKP4fVAvI&t=120s}
\BIBentrySTDinterwordspacing

\bibitem{RaveshSNS}
Y.~Li, R.~Sukhnandan, J.~P. Gill, H.~J. Chiel, V.~Webster-Wood, and R.~D. Quinn, ``A bioinspired synthetic nervous system controller for pick-and-place manipulation,'' in \emph{2023 IEEE International Conference on Robotics and Automation (ICRA)}, 2023, pp. 8047--8053.

\bibitem{vickieSNS}
Y.~Li, V.~A. Webster-Wood, J.~P. Gill, G.~P. Sutton, H.~J. Chiel, and R.~D. Quinn, ``A computational neural model that incorporates both intrinsic dynamics and sensory feedback in the aplysia feeding network,'' \emph{Biological Cybernetics}, pp. 1--27, 2024.

\bibitem{Naughton2020Elastica:Control}
N.~Naughton, J.~Sun, A.~Tekinalp, T.~Parthasarathy, G.~Chowdhary, and M.~Gazzola, ``Elastica: A compliant mechanics environment for soft robotic control,'' \emph{IEEE Robotics and Automation Letters}, vol.~6, no.~2, pp. 3389--3396, 2021.

\bibitem{ppo_paper}
\BIBentryALTinterwordspacing
J.~Schulman, F.~Wolski, P.~Dhariwal, A.~Radford, and O.~Klimov, ``Proximal policy optimization algorithms,'' \emph{CoRR}, vol. abs/1707.06347, 2017. [Online]. Available: \url{http://arxiv.org/abs/1707.06347}
\BIBentrySTDinterwordspacing

\bibitem{schaffer2023tall}
S.~Schaffer and V.~A. Webster-Wood, ``{The Tall, the Squat, \& the Bendy: Parametric Modeling and Simulation Towards Multi-functional Biohybrid Robots},'' \emph{Biomimetic and Biohybrid Systems}, pp. 217--226, 2023.

\bibitem{Gazzola2018ForwardFilaments}
M.~Gazzola, L.~H. Dudte, A.~G. McCormick, and L.~Mahadevan, ``{Forward and inverse problems in the mechanics of soft filaments},'' \emph{Royal Society Open Science}, vol.~5, no.~6, 6 2018.

\bibitem{cosserat1909theorie}
E.~M.~P. Cosserat and F.~Cosserat, \emph{Th{\'e}orie des corps d{\'e}formables}.\hskip 1em plus 0.5em minus 0.4em\relax A. Hermann et fils, 1909.

\bibitem{kim2021review}
D.~Kim, S.-H. Kim, T.~Kim, B.~B. Kang, M.~Lee, W.~Park, S.~Ku, D.~Kim, J.~Kwon, H.~Lee \emph{et~al.}, ``Review of machine learning methods in soft robotics,'' \emph{Plos one}, vol.~16, no.~2, p. e0246102, 2021.

\bibitem{truly_ppo}
\BIBentryALTinterwordspacing
Y.~Wang, H.~He, and X.~Tan, ``Truly proximal policy optimization,'' in \emph{Proceedings of The 35th Uncertainty in Artificial Intelligence Conference}, ser. Proceedings of Machine Learning Research, R.~P. Adams and V.~Gogate, Eds., vol. 115.\hskip 1em plus 0.5em minus 0.4em\relax PMLR, 22--25 Jul 2020, pp. 113--122. [Online]. Available: \url{https://proceedings.mlr.press/v115/wang20b.html}
\BIBentrySTDinterwordspacing

\bibitem{stable-baselines3}
\BIBentryALTinterwordspacing
A.~Raffin, A.~Hill, A.~Gleave, A.~Kanervisto, M.~Ernestus, and N.~Dormann, ``Stable-baselines3: Reliable reinforcement learning implementations,'' \emph{Journal of Machine Learning Research}, vol.~22, no. 268, pp. 1--8, 2021. [Online]. Available: \url{http://jmlr.org/papers/v22/20-1364.html}
\BIBentrySTDinterwordspacing

\bibitem{towers_gymnasium_2023}
\BIBentryALTinterwordspacing
M.~Towers, J.~K. Terry, A.~Kwiatkowski, J.~U. Balis, G.~d. Cola, T.~Deleu, M.~Goulão, A.~Kallinteris, A.~KG, M.~Krimmel, R.~Perez-Vicente, A.~Pierré, S.~Schulhoff, J.~J. Tai, A.~T.~J. Shen, and O.~G. Younis, ``Gymnasium,'' Mar. 2023. [Online]. Available: \url{https://zenodo.org/record/8127025}
\BIBentrySTDinterwordspacing

\bibitem{brockman2016openai}
G.~Brockman, V.~Cheung, L.~Pettersson, J.~Schneider, J.~Schulman, J.~Tang, and W.~Zaremba, ``{OpenAI Gym},'' \emph{arXiv preprint arXiv:1606.01540}, 2016.

\bibitem{stable_baselines3_ppo}
\BIBentryALTinterwordspacing
{Stable-Baselines3 Contributors}, ``{PPO - Proximal Policy Optimization},'' 2024, accessed: 2024-07-23. [Online]. Available: \url{https://stable-baselines3.readthedocs.io/en/master/modules/ppo.html}
\BIBentrySTDinterwordspacing

\bibitem{bridges2}
S.~T. Brown, P.~Buitrago, E.~Hanna, S.~Sanielevici, R.~Scibek, and N.~A. Nystrom, ``Bridges-2: A platform for rapidly-evolving and data intensive research,'' in \emph{Practice and Experience in Advanced Research Computing}, 2021, pp. 1--4.

\bibitem{curriculum-learning}
S.~Narvekar, B.~Peng, M.~Leonetti, J.~Sinapov, M.~E. Taylor, and P.~Stone, ``Curriculum learning for reinforcement learning domains: A framework and survey,'' \emph{Journal of Machine Learning Research}, vol.~21, no. 181, pp. 1--50, 2020.

\end{thebibliography}

\end{document}